\documentclass[runningheads]{llncs}

\usepackage{eccv}

\usepackage{eccvabbrv}

\usepackage{graphicx}
\usepackage{booktabs}
\usepackage{multirow}

\usepackage[accsupp]{axessibility}  

\usepackage{hyperref}
\hypersetup{
    colorlinks=true,    
    citecolor=green,    
    urlcolor=red        
}

\usepackage{orcidlink}

\begin{document}

\title{IFTR: An Instance-Level Fusion Transformer for Visual Collaborative Perception} 

\titlerunning{IFTR: An Instance-Level Fusion Transformer}

\author{Shaohong Wang\inst{1} \and
Lu Bin\inst{1} \and
Xinyu Xiao\inst{2} \and
Zhiyu Xiang\inst{1} \and
Hangguan Shan\inst{1} \and
Eryun Liu\inst{1}\thanks{Corresponding author. }}

\authorrunning{S.~Wang et al.}

\institute{Zhejiang University, Hangzhou, China \\
\email{\{wangsh0111,lubin2022, xiangzy, hshan, eryunliu\}@zju.edu.cn} \and
Institute of Automation of Chinese Academy of Sciences, Beijing, China\\
\email{xinyu.xiao@nlpr.ia.ac.cn}}

\maketitle

\begin{abstract}
 Multi-agent collaborative perception has emerged as a widely recognized technology in the field of autonomous driving in recent years. However, current collaborative perception predominantly relies on LiDAR point clouds, with significantly less attention given to methods using camera images. This severely impedes the development of budget-constrained collaborative systems and the exploitation of the advantages offered by the camera modality. This work proposes an instance-level fusion transformer for visual collaborative perception (IFTR), which enhances the detection performance of camera-only collaborative perception systems through the communication and sharing of visual features. To capture the visual information from multiple agents, we design an instance feature aggregation that interacts with the visual features of individual agents using predefined grid-shaped bird eye view (BEV) queries, generating more comprehensive and accurate BEV features. Additionally, we devise a cross-domain query adaptation as a heuristic to fuse 2D priors, implicitly encoding the candidate positions of targets. Furthermore, IFTR optimizes communication efficiency by sending instance-level features, achieving an optimal performance-bandwidth trade-off. We evaluate the proposed IFTR on a real dataset, DAIR-V2X, and two simulated datasets, OPV2V and V2XSet, achieving performance improvements of 57.96\%, 9.23\% and 12.99\% in AP@70 metrics compared to the previous SOTAs, respectively. Extensive experiments demonstrate the superiority of IFTR and the effectiveness of its key components. The code is available at \href{https://github.com/wangsh0111/IFTR}{https://github.com/wangsh0111/IFTR}.

\end{abstract}

\section{Introduction}
\label{sec:intro}
3D object detection is a fundamental task in computer vision, aiming to locate objects in 3D physical space given sensor inputs. It is crucial in various applications such as autonomous driving \cite{li2022bevformer, zhang2023monodetr, qi2017pointnet,yan2018second}, robotics \cite{li2022dual}, and unmanned aerial vehicles \cite{hu2023aerial}. Despite significant advancements in LiDAR-based methods \cite{lang2019pointpillars, zhou2018voxelnet,yin2021center,deng2021voxel}, camera-based approaches \cite{wang2021fcos3d, philion2020lift,wang2022detr3d,reading2021categorical,yin2021center} have garnered widespread attention in recent years. In comparison to LiDAR, cameras have lower deployment costs, are more easily scalable, and provide valuable visual cues for detecting distant objects and identifying vision-based road elements (such as traffic lights and lane markings) \cite{li2022bevformer}. However, due to the lack of depth information in RGB images, solely camera-based 3D object detection is often noticeably inferior to lidar-based detection in most scenarios.

The latest advancements in V2X communication technology and Intelligent Transportation Systems enable intelligent agents to share information between each other \cite{betz2022autonomous,shladover2021opportunities,raboy2021proof,guo2020evaluating,yu2022dair}. Through multi-agent interactions, the limitations of individual agent perception, such as occlusion and long-range issues, can be alleviated, thereby enhancing the performance of camera-based 3D object detection. Currently, the prevailing frameworks for collaborative perception \cite{wang2020v2vnet,xu2022opv2v,xu2022v2x,hu2022where2comm,xu2022cobevt} primarily involve the fusion at the Bird's Eye View (BEV) feature level. However, this paradigm is highly sensitive to the accuracy of BEV features for individual agents and faces challenges in spatial alignment, imposing stringent requirements on real-time poses between agents, leading to insufficient robustness in feature fusion. Nevertheless, due to the lack of depth information in RGB images, generating BEV features from a 2D plane is ill-posed. Therefore, the detection performance of visual collaborative perception methods based solely on BEV features level in camera settings is impacted by compounded errors, where inaccurate BEV features can severely impact the final performance.

Diverging from multi-camera 3D object detection, V2X collaborative perception must also account for the constraints of communication bandwidth. It is crucial to address how to reduce communication bandwidth while ensuring collaborative perception performance. Each agent must select visual features with the maximum information content to optimize communication efficiency. 

In this paper, we propose IFTR, a robust and efficient transformer-based instance-level feature fusion framework. Our central idea is to extract spatial features from regions of interest in cross-agent camera views to generate BEV features, and implicitly encode candidate positions of objects in 3D space through a heuristic adaptive approach. Specifically, IFTR incorporates three key designs: i) Instance feature aggregation, aggregating spatial information from visual features of multiple agent camera views using predefined grid-shaped BEV queries. This is achieved through a transformer to implement robust and adaptive feature fusion, avoiding strict spatial alignment requirements in BEV feature-level fusion strategies. It also eliminates the dependency on high-accuracy single-agent BEV features and real-time pose accuracy. ii) Message selection and feature map reconstruction, reducing bandwidth consumption by sharing instance-level visual features rather than entire image features. iii) Cross-domain query adaptation, encoding instance-level features as object queries, combining them with the original learnable object queries to form the new hybrid object queries. This is then input into a deformable DETR head \cite{carion2020end,zhu2020deformable} for end-to-end 3D object detection. 

To evaluate IFTR, we conduct extensive experiments on a real-world dataset, DAIR-V2X \cite{yu2022dair}, and two simulated datasets, OPV2V \cite{xu2022opv2v} and V2XSet \cite{xu2022v2x}. The experimental results indicate that IFTR significantly outperforms previous works in the performance-bandwidth trade-off across multiple datasets. To sum up, the main contributions are as follows:
\begin{itemize}
    \item We introduce IFTR, a superior camera-only instance-level feature fusion framework based on transformers. This framework addresses occlusion and long-range issues through multi-agent collaboration, achieving more accurate and complete 3D object detection.
    \item We devise an instance feature aggregation that utilizes learnable BEV queries to aggregate spatial information from multiple agents, resulting in more accurate BEV features.
    \item We design a cross-domain query adaptation module by encoding instance-level features as object queries, facilitating the implicit encoding of candidate positions for targets in 3D space, which alleviates the problem of target spatial distribution differences between the training set and the test set.
    \item We develope communication-efficient collaboration techniques by sharing the most informative visual features to reduce communication overhead. Our IFTR achieves state-of-the-art performance on a real world dataset, DAIR-V2X, and two simulated datasets, OPV2V and V2XSet, with AP@70 metrics improving by 57.96\%, 9.23\% and 12.99\% compared to the previous SOTAs, respectively.
\end{itemize}

\section{Related Work}
\subsection{Camera-based 3D Object Detection}
Given an RGB image and the corresponding camera parameters, the purpose of image-based 3D object detection is to classify and locate the objects of interest. Since images lack depth information, this problem is ill-posed and more challenging compared to 2D detection. FCOS3D \cite{wang2021fcos3d} and SMOKE \cite{liu2020smoke} extend 2D detection networks by utilizing fully convolutional networks to directly regress the depth for each object. Pseudo LiDAR \cite{wang2019pseudo} employs a depth estimation network to transform 2D images into 3D pseudo-point cloud signals, followed by the use of a lidar-based detection network for 3D detection. DETR3D \cite{wang2022detr3d} utilizes a set of sparse 3D queries to index 2D features extracted from multi-view images, samples corresponding features, enabling end-to-end 3D detection box prediction without the need for time-consuming post-processing like NMS (Non-Maximum Suppression).

Recently, there has been widespread attention on BEV representations due to their ability to clearly depict object positions and scales, coupled with efficient computational efficiency. OFT \cite{roddick2018orthographic} and ImVoxelNet \cite{rukhovich2022imvoxelnet} project predefined voxels onto image features, generating voxel representations of the scene. LSS \cite{philion2020lift} and CaDDN \cite{reading2021categorical} utilize view transformation modules to convert dense 2D image features into BEV space. CVT \cite{zhou2022cross} leverages position encoding from camera perception and dense cross-attention to correlate perspective view and BEV view features. BEVDet \cite{huang2021bevdet,huang2022bevdet4d} and BEVDepth \cite{li2023bevdepth} use lift-splat operations for view transformation. PETR \cite{liu2022petr,liu2023petrv2} utilizes 3D position encoding and global cross-attention for feature fusion, transforming 2D features into 3D position-aware representations. BEVFormer \cite{li2022bevformer,yang2023bevformer} effectively aggregates spatio-temporal features from a surround-view camera and historical BEV features using deformable attention, generating unified BEV features. SparseBEV \cite{liu2023sparsebev} adapts self-attention with an adaptive receptive field for feature aggregation in BEV space, enjoying the advantages of BEV space without explicitly constructing dense BEV features. In this work, we choose the single-scale and history-free BEVFormer-S \cite{li2022bevformer} as the single-agent 3D object detector for computational efficiency.

\subsection{Collaborative Perception}
Collaborative perception is an emerging application in multi-agent systems, aiming to enhance perceptual performance by facilitating communication among other agents for the shared exchange of information. Based on its message sharing strategy, it can be categorized into three types: early fusion (i.e. sharing raw sensor information), intermediate fusion (i.e. sharing intermediate layer features of neural networks), and late fusion (i.e. sharing 3D detection results, such as 3D bounding box position and confidence score). Early fusion usually requires large transmission bandwidth, while late fusion cannot provide valuable scenario context. Intermediate fusion achieves a better trade-off between performance and bandwidth. When2com \cite{liu2020when2com} introduces a handshake mechanism to determine when to communicate and create sparse communication graph. OPV2V \cite{xu2022opv2v} utilizes a single-head self-attention module to fuse features, while F-Cooper \cite{chen2019f} employs a maxout \cite{goodfellow2013maxout} fusion operation. V2VNet \cite{wang2020v2vnet} utilizes a spatially aware graph neural network to aggregate shared feature representations among multiple vehicles. DiscoNet \cite{li2021learning} adopts knowledge distillation to leverage the advantages of both early and intermediate collaboration. V2X-ViT \cite{xu2022v2x} introduces a novel heterogeneous multi-agent attention module to fuse information across heterogeneous agents. Where2comm \cite{hu2022where2comm} introduces a spatial confidence map to reduce communication bandwidth consumption. TransIFF \cite{chen2023transiff} further reduces communication bandwidth consumption by transmitting object queries. CoAlign \cite{lu2023robust} employs learnable or mathematical methods to correct pose errors for more accurate collaborative perception. HM-ViT \cite{xiang2023hm} and HEAL \cite{lu2024extensible} investigate multi-agent heterogeneous modality collaborative perception problems with different sensor modalities, expanding the scale of collaboration.

However, prior collaborative perception efforts \cite{xu2022opv2v,wang2020v2vnet,xu2022v2x,hu2022where2comm,chen2023transiff,lu2023robust} have primarily focused on lidar-based 3D object detection, yielding suboptimal performance in a camera-only setup. We speculate that the reason lies in the inferior quality of BEV features from a single-agent compared to lidar-based methods, resulting in suboptimal solutions at the BEV feature level. In this paper, we propose a novel camera-only instance-level feature fusion framework with lower communication bandwidth consumption. It generates more accurate BEV features and 3D object queries tailored to each instance by leveraging instance-level features shared among multiple agents, thereby achieving more efficient and practical collaborative perception.

\section{IFTR}
\label{sec:iftr}

\begin{figure}[tb]
  \centering
  \includegraphics[width=12cm]{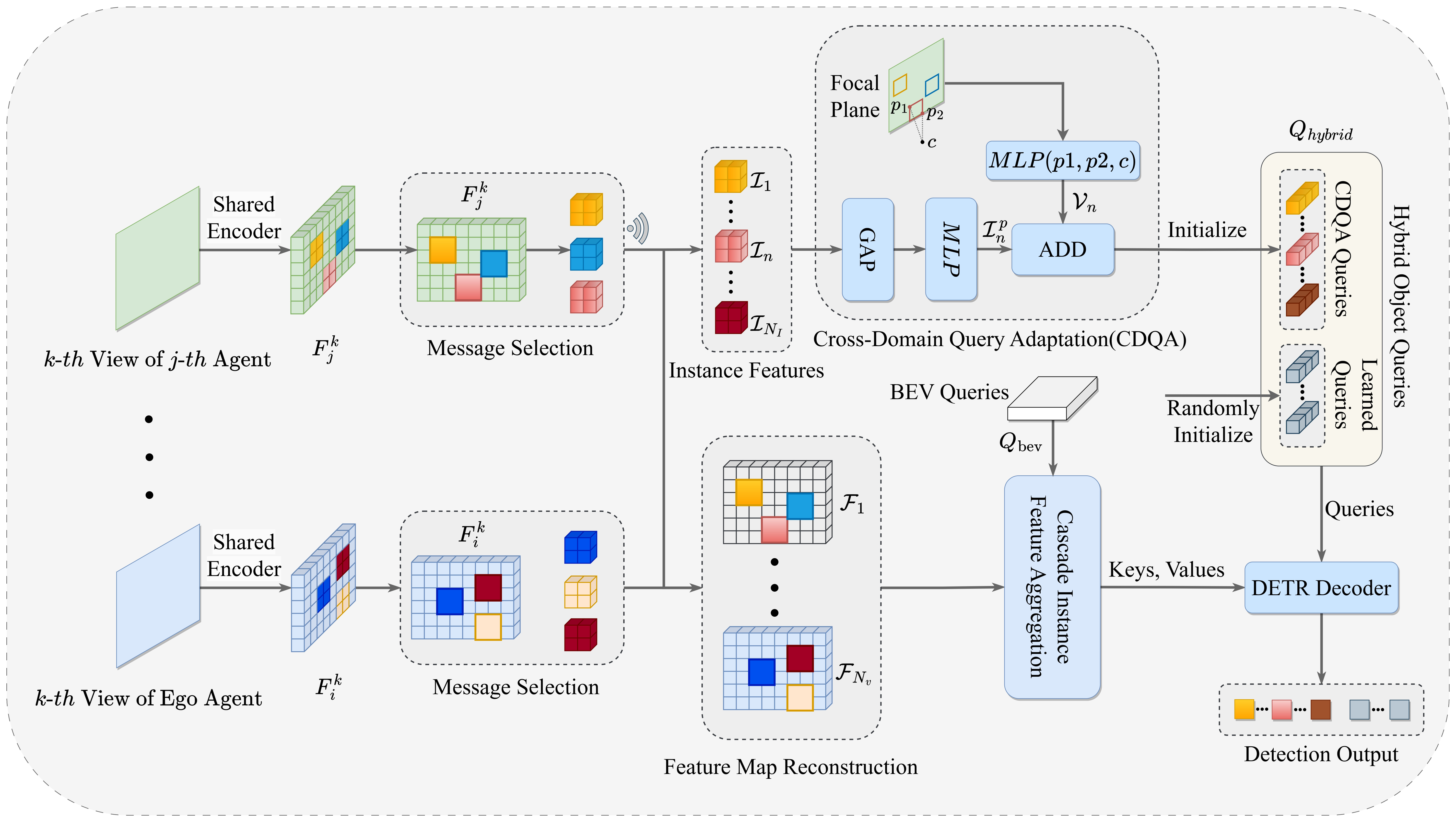}
  \caption{
  Overall architecture of IFTR. i) We employ the message selection and feature map reconstruction module to share instance-level features, reducing bandwidth consumption; ii) In instance feature aggregation (IFA), each BEV query interacts only with image features from regions of interest from multiple views; iii) In CDQA, we encode the feature map information and 3D positional information of each instance into 3D object query.
  }
  \label{fig:iftr}
\end{figure}

This section presents IFTR, a robust and efficient camera-only instance-level feature fusion framework for collaborative perception. As depicted in \cref{fig:iftr}, IFTR comprises an image encoder, message selection and feature map reconstruction, cascade instance feature aggregation, cross-domain query adaptation, and a DETR decoder. The image encoder is utilized to extract visual features from the input images. The proposed message selection and feature map reconstruction module are employed at the sender and receiver ends, respectively, for instance-level visual feature filtering and image feature reconstruction, effectively saving communication bandwidth. Based on visual features from multiple camera views, the proposed cascade instance feature aggregation generates comprehensive and accurate BEV features; the proposed cross-domain query adaptation module encodes instance-level visual features into object queries to encode candidate positions for the targets, combining them with the original learnable object queries to form the hybrid object queries.  Ultimately, the DETR decoder takes the BEV features and the hybrid object queries as inputs to predict the target's category and location information.

\subsection{Image Encoder}
We leverage EfficientNet \cite{tan2019efficientnet,tan2021efficientnetv2} to extract visual features from the input images because of its high performance and low computational complexity. Given $L$ agents and $N_{cams}$ camera views for each agent, $F_i^k\in \mathbb{R}^{fC \times f H \times f W}$ is used to represent the extracted feature map of the $k$-th camera view of the $i$-th agent, where $fC$, $fH$ and $fW$ are the channel, height and width of the feature map. Considering different agents may have different camera views, the feature maps of views that not exist are set to zero when the actual number of camera views less than $N_{\text{cams}}$ and use a mask to indicate the valid feature maps.

\subsection{Message Selection and Feature Map Reconstruction}
To achieve accurate and comprehensive 3D object detection, it is imperative for each agent to exchange visual features and leverage complementary information. We propose communication-efficient collaborative techniques by sharing the most informative visual features to reduce communication overhead. The intuition is that foreground regions contain more informative content than background regions. During collaboration, visual features of regions with objects can assist in recovering missed objects due to occlusion or long-range issues, while background regions can be omitted to reduce the system's communication load. 

Given a feature map $F_j^k$, the message sent for sharing is selected by the following steps:
\begin{enumerate}
    \item Extracting a set of candidate object instance by a 2D object detector. All the confidence values of instance are greater than a predefined threshold, $c_{thre}$.
    \item Cropping the feature map of each instance by the predicted bounding box in inference stage or by the ground truth bounding box in the training stage.
    \item The cropped feature map of the $n$-th selected instance, ${F}_{j\to i}^{k,n}$, and bounding box are sent to the $i$-th agent for sharing. 
\end{enumerate}

At the side of ego agent (or the $i$-th agent), the received object instance feature maps of the $k$-th camera view of the $j$-th agent are first combined together to reconstruct the feature map, $\mathcal{F}_{j \to i}^{k}$,  of foreground regions of original feature map $F_j^k$ by filling back the cropped features. The features in background region are set to zero. 

\subsection{Instance Feature Aggregation}
\label{sec:ifa}

\begin{figure}[tb]
  \centering
  \includegraphics[width=12cm]{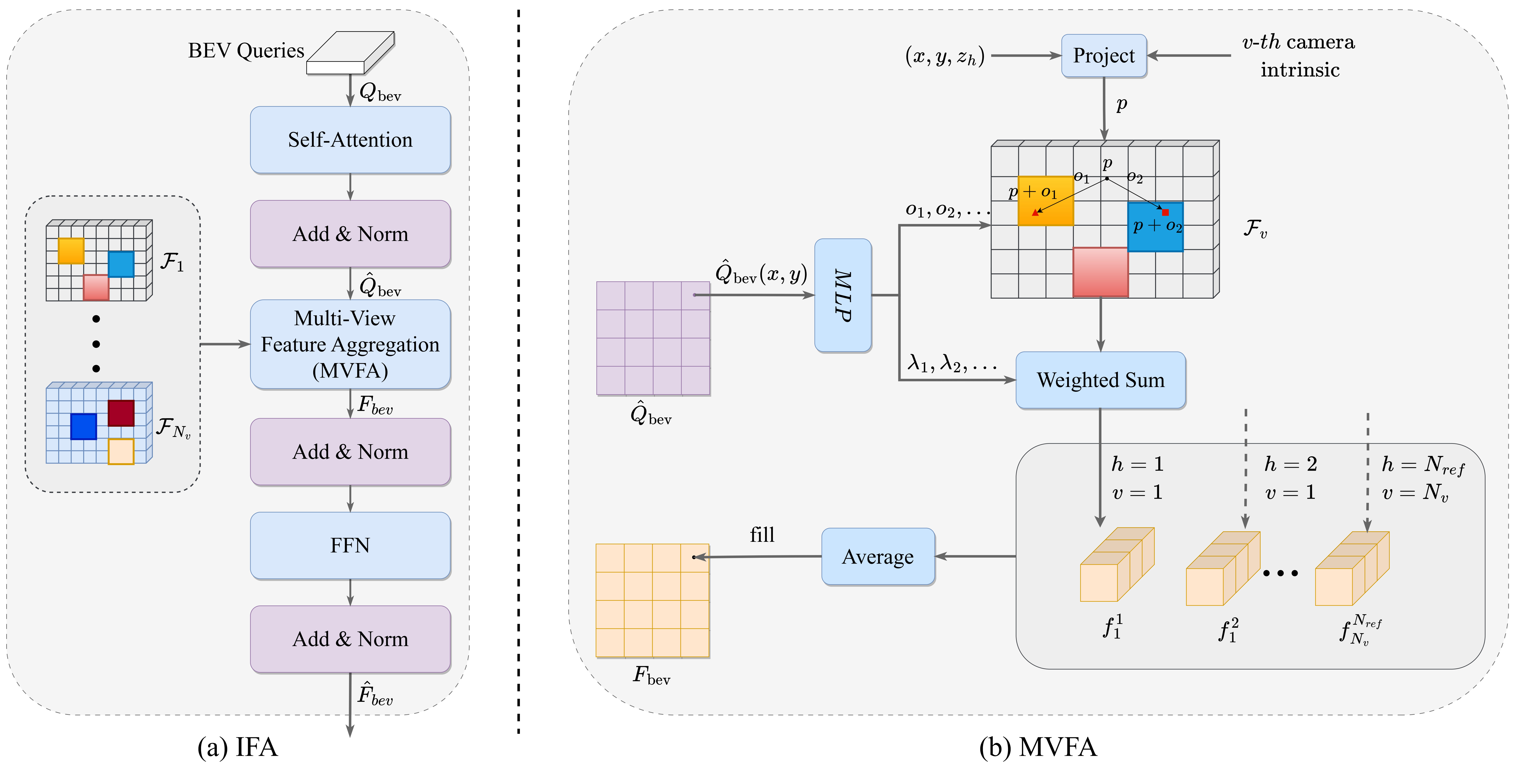}
  \caption{
  (a) The architecture of our proposed Instance Feature Aggregation (IFA);
  (b) Multi-View Feature Aggregation (MVFA) illustrated in \cref{sec:ifa}
  }
  \label{fig:ifa}
\end{figure}

Given a set of received foreground feature maps, $\{\mathcal{F}_{j\to i}^{k}| j=1,2,\dots,N_{agent}, k=1,2,\dots,N_{cams}\}$, where $N_{agent}$ is the number of agents, the purpose of instance feature aggregation (IFA) module is to form a BEV feature map by aggregating features from all received foreground feature maps. Since all camera views are treated equally in IFA module, for simplicity, we represent all the foreground feature maps by $\{\mathcal{F}_{v}| v=1,2,\dots,N_v\}$, where the total number of views is $N_v = N_{agent}\times N_{cams}$.

The BEV space around the ego agent is divided into a discrete grid space of size $H\times W$. The origin of ego coordinate is at the center of BEV grid. To construct a feature map on the BEV space of the ego agent by fusing reconstructed feature maps $\{\mathcal{F}_i|i=1,2,\dots, N_v\}$ from all collaborative views, we propose a deformable attention based cascade instance feature aggregation (IFA) module (see \cref{fig:ifa}). The BEV feature map is obtained by the following steps:
\begin{enumerate}
    \item A set of learnable parameters $Q_{\text{bev}} \in \mathbb{R}^{C \times H \times W}$ is predefined and randomly initialized at the first IFA block.
    \item For a given grid centered at $(x, y)$, a set of points $\{(x, y, z_h) | h=1, 2, \dots, N_{ref}\}$ is selected as reference points, where $\{z_h|h=1,2,\dots, N_{ref}\}$ are predefined heights for each grid. The reference points are used as feature indices to gather information from collaborative feature maps.
    \item For a reference point $(x, y, z_h)$ at grid $(x, y)$, the corresponding feature vector is aggregated by
    $$
    F_{bev}^h(x, y) =\frac{1}{|w|} \sum_{v=1}^{N_{v}} w_v f_v^h,
    $$
    where $w_v\in \{0,1\}$ indicates whether the reference point $(x, y, z_h)$ is projected inside the image space of the $v$-th camera view, and $|w| = \sum_v w_v$ is total number of views observing the reference point, and $f_v^h$ is the output of a deformable attention module, which will be described later in this section.
    \item The aggregated BEV feature $F_{bev}$ is the average feature over all $N_{ref}$ reference points for each grid and followed by a feed forward network (FFN) and an Add\&Norm module to get the output $\hat{{F}}_{bev}$ of IFA module.
    \item The above four steps assemble an IFA module. To sufficiently fuse all camera view features, we cascade 6 IFA modules sequentially, with the previous IFA output $\hat{{F}}_{bev}$ as the input $Q_{\text{bev}}$ of the next IFA.
\end{enumerate}

Given a reference point $(x, y, z_h)$ at grid $(x, y)$ and the feature map $\mathcal{F}_{v}$ of $v-$th camera view, if point $(x, y, z_h)$ is observed by the $v$-th camera view, the deformable attention first takes a feature vector $\hat{Q}_{\text{bev}}(x, y)$ from $\hat{Q}_{\text{bev}}$ and the vector $\hat{Q}_{\text{bev}}(x, y)$ is then fed into a MLP layer. The MLP layer output a list of offset vectors $(o_1, o_2, \dots, o_{N_{da}}\}$ and a weight vector $\lambda=(\lambda_1, \lambda_2, \dots, \lambda_{N_{da}})$, where $N_{da}$ is the number of positions in attention, the output of deformable attention is
$$
f_v^h =\sum_{i=1}^{N_{da}} \lambda_i\mathcal{F}_{v}(p + o_i),
$$
where $p$ is the projected image coordinate of $(x, y, z_h)$ on $v$-th view.

\subsection{Cross-Domain Query Adaptation}
The object decoder of BEVformer \cite{li2022bevformer} is a DETR \cite{carion2020end,zhu2020deformable} decoder that receives a set of learned embeddings as object queries, aiming to learn potential locations of target objects during training. However, it takes a considerable amount of time for randomly initialized embeddings to converge to suitable positions. Additionally, during inference, object queries remain fixed for all images, while the spatial distribution of target objects may vary, leading to potentially inaccurate results. To address this issue, we employ a cross-domain query adaptation(CDQA) module to initialize object queries for decoder. This module implicitly encodes candidate positions for the target, making it easier for the detection head to capture the target object accurately.

The maximum number of queries for decoder is set to $N_{Q}$. The initial set of queries consists of two parts. The first part of queries is obtained from selected instances of all ego and collaborative camera views and the second part (the rest) of queries is randomly initialized.

Given a set of shared instance features $\{(\mathcal{I}_{n}, B_n) | n=1,2,\dots, N_I\}$ from all collaborative and ego views, where $N_I$ is total number of instances, $\mathcal{I}_{n}$ is the $n$-th cropped instance feature map, and $B_n$ is the bounding box coordinates, the queries is obtained as following
\begin{enumerate}
    \item Global average pooling (GAP) on instance feature maps and then followed by a MLP module to get a set of feature vector $\{\mathcal{I}_{n}^p | n=1,2,\dots, N_I\}$.
    \item Encode the instance position information into a vector by a MLP layer, obtaining a set of position embeddings $\{\mathcal{V}_{n} | n=1,2,\dots, N_I\}$. The input to the MLP is the concatenate of three coordinates $(p_1, p_2, c)$, where $p_1, p_2$ are the 3D coordinates of left-top and bottom-right point of instance bounding box on the focal plane and $c$ is the coordinate of corresponding camera origin. All coordinates are projected to the ego coordinate system. The three points indicate a cone where the object lies.
    \item Combine the feature vectors and position embeddings by adding, obtaining the instance feature initialized set of queries $\{\mathcal{I}_n^p + \mathcal{V}_{n} | n=1,2,\dots, N_I\}$ for object decoder.
\end{enumerate}
The union, $Q_{\text{hybrid}}$, of the instance feature initialized set of queries and the randomly initialized set of queries is then used as queries of object decoder.

\subsection{Object Decoder}
The object decoder of IFTR employs a deformable DETR detection head, taking the BEV features \( \hat{{F}}_{bev} \in \mathbb{R}^{C \times H \times W} \) and the hybrid object queries \( Q_{\text{hybrid}} \in \mathbb{R}^{N_Q \times C} \) as input, producing detection results \( \mathcal{O} = \Phi_{\text{dec}}(\hat{{F}}_{bev}, Q_{\text{hybrid}}) \in \mathbb{R}^{N_Q \times 8} \). Each position of \( \mathcal{O} \) represents a rotated box with class \((c, x, y, z, w, l, h, \theta)\), denoting class confidence, position, size and angle. The objects are the final output of the proposed collaborative perception system. We utilize the $\ell_{1}$ loss for regression and the focal loss \cite{lin2017focal} for classification.

\section{Experiments}
\subsection{Experimental Setup}
\subsubsection{Datasets.} 
To evaluate the performance of IFTR on the visual collaborative perception task, we conduct extensive experiments on three multi-agent datasets, including V2XSet \cite{xu2022v2x}, OPV2V \cite{xu2022opv2v}, and DAIR-V2X \cite{yu2022dair}. V2XSet \cite{xu2022v2x} is a simulated dataset supporting V2X perception, co-simulated by CARLA \cite{dosovitskiy2017carla} and OpenCDA \cite{xu2021opencda}. It comprises 73 representative scenes with 2 to 5 connected agents and 11,447 annotated 3D frames, with an image resolution of $600 \times 800$. OPV2V \cite{xu2022opv2v} is a large-scale vehicle-to-vehicle collaborative perception dataset, comprising 12,000 LiDAR point cloud frames and RGB images with 230,000 annotations of 3D bounding boxes, with an image resolution of $600 \times 800$. DAIR-V2X \cite{yu2022dair} is a real-world vehicle-to-infrastructure collaborative perception dataset, consists of 100 realistic scenes and 18,000 data samples, with an image resolution of $1080 \times 1920$.

\subsubsection{Implementation details.}
We use the single-scale and history-free BEVFormer-S \cite{li2022bevformer} as the 3D object detector for single agent. We employ EfficientNet \cite{tan2019efficientnet,tan2021efficientnetv2} as the image backbone for better computation efficiency and utilize a smaller grid resolution (0.4 m) to retain fine-trained spatial details. The intermediate BEV feature map has a dimension of $256 \times 256 \times 256$, with a perception range defined as $x \in [-51.2 \, \text{m}, 51.2 \, \text{m}]$ and $y \in [-51.2 \, \text{m}, 51.2 \, \text{m}]$. Following prior research \cite{xu2022opv2v,xu2022v2x,xu2022cobevt}, we only modified the fusion module of different intermediate fusion methods while keeping the other components such as camera feature extractor and detection head the same. As IFTR involves instance-level fusion of 2D image features, we construct it following the method proposed in \cref{sec:iftr} and straightforwardly add a late fusion strategy to explore its combined effects with other fusion strategies.

\subsubsection{Evaluation metrics.}
We evaluate the detection performance using Average Precision (AP) metrics with Intersection over Union (IoU) thresholds of 0.30, 0.50, and 0.70. Additionally, the communication volume follows the standard setting as \cite{xu2022opv2v,xu2022v2x,hu2022where2comm,chen2023transiff} that counts the message size by byte in log scale with base 2.

\subsection{Quantitative Evaluation}
\subsubsection{Benchmark comparison.}

\begin{table}[tb]
  \caption{
    3D detection performance comparison on  the V2XSet \cite{xu2022v2x}, OPV2V \cite{xu2022opv2v} and DAIR-V2X \cite{yu2022dair} datasets. The results are reported in Average Precision (AP) at IoU=0.30, 0.50 and 0.70 on perfect settings.
  }
  \label{tab:performance_metrics}
  \centering
  \begin{tabular}{@{}c|ccc|ccc|ccc@{}}
    \hline
        \multirow{2}{*}{Method} & 
        \multicolumn{3}{c|}{V2XSet} & \multicolumn{3}{c|}{OPV2V} & \multicolumn{3}{c}{DAIR-V2X} \\
         & AP@30   & AP@50   & AP@70   & AP@30   & AP@50   & AP@70 
         & AP@30    & AP@50   & AP@70 \\ \hline
        No Collaboration 
        & 0.4343 & 0.3037 & 0.1379 & 0.5652 & 0.4594 & 0.2556 & 0.2102 & 0.0709 & 0.0080 \\
        Late Fusion      
        & 0.6803 & 0.5141 & 0.2559 & 0.8789 & 0.7762 & 0.5192 & 0.2952 & 0.1631 & 0.0421 \\
        V2VNet \cite{wang2020v2vnet}           
        & 0.6886 & 0.5954 & 0.3900 & 0.8682 & 0.7906 & 0.5759 & 0.2581 & 0.1354 & 0.0276 \\
        V2X-ViT \cite{xu2022v2x}          
        & 0.6690 & 0.5914 & 0.4123 & 0.8512 & 0.7841 & 0.5838 & 0.2544 & 0.1402 & 0.0277 \\
        Where2comm \cite{hu2022where2comm}       
        & 0.6969 & 0.6169 & 0.4396 & 0.8471 & 0.7714 & 0.5860 & 0.2613 & 0.1424 & 0.0296 \\
        CoBEVT \cite{xu2022cobevt}           
        & 0.6629 & 0.5884 & 0.4081 & 0.8709 & 0.8026 & 0.5934 & 0.2608 & 0.1453 & 0.0255 \\
        CoAlign \cite{lu2023robust}          
        & 0.7564 & 0.6479 & 0.3964 & 0.8689 & 0.8021 & 0.6046 & 0.2645 & 0.1470 & 0.0262 \\
        IFTR             & 
        \textbf{0.8057} & \textbf{0.7173} & \textbf{0.4967} &
        \textbf{0.9233} & \textbf{0.8556} & \textbf{0.6604} &
        \textbf{0.3855} & \textbf{0.2058} & \textbf{0.0665} \\ \hline
  \end{tabular}
\end{table}

\begin{figure}[tb]
    \centering
    \label{fig:agent_num}
    \begin{subfigure}{0.45\textwidth}
        \centering
        \includegraphics[width=\linewidth]{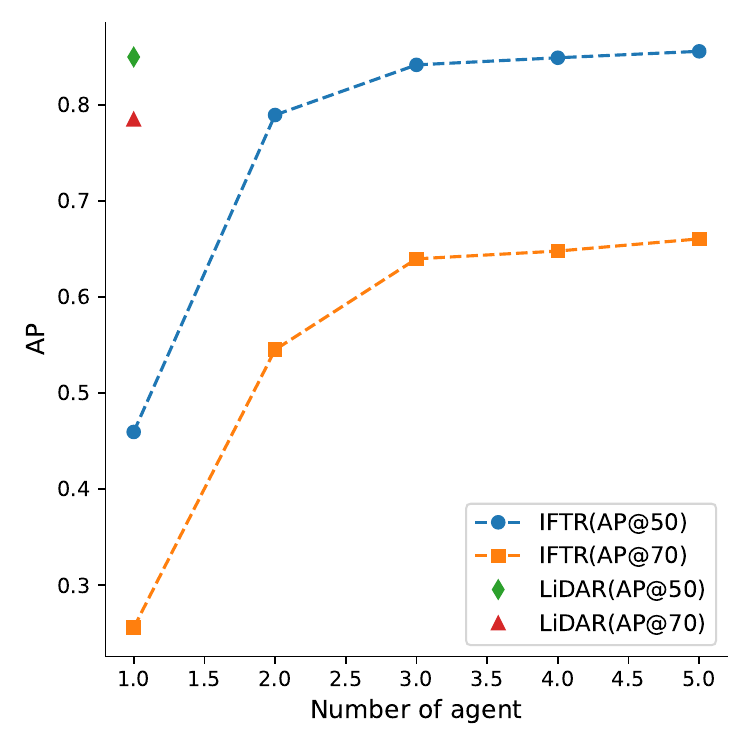}
        \caption{All scenes on OPV2V}
        \label{fig:agent_all}
    \end{subfigure}
    \hfil
    \begin{subfigure}{0.45\textwidth}
        \centering
        \includegraphics[width=\linewidth]{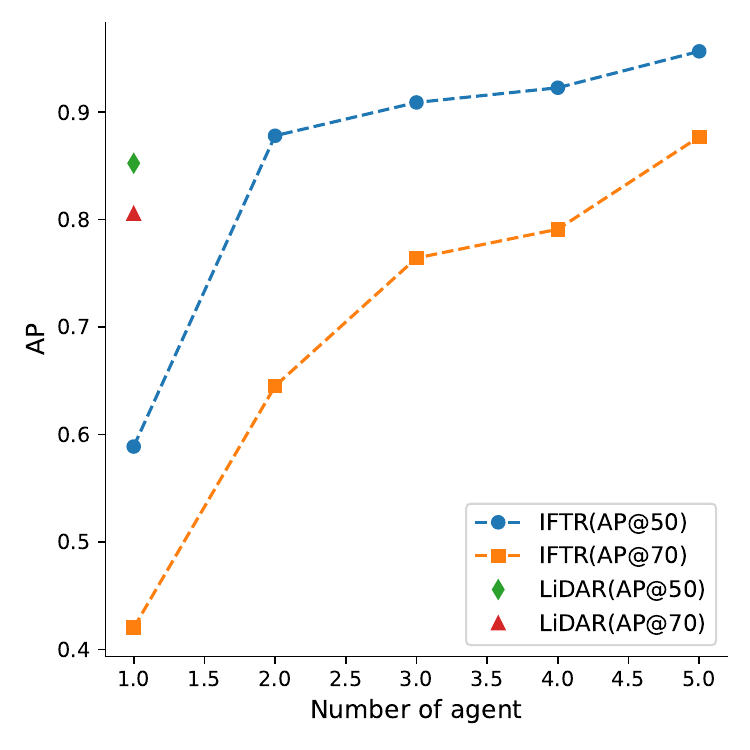}
        \caption{A certain scene on OPV2V}
        \label{fig:agent_certain}
    \end{subfigure}
    \caption{
    IFTR steadily improves 3D detection performance as the number of agents grows.
    (a) The relationship between 3D detection performance and the maximum collaboration count on the OPV2V test set; 
    (b) The relationship between 3D detection performance and collaboration count in a certain scene on the OPV2V test set.
    }
\end{figure}

\begin{figure}[tb]
    \centering
    \begin{subfigure}{0.45\textwidth}
        \centering
        \includegraphics[width=\linewidth]{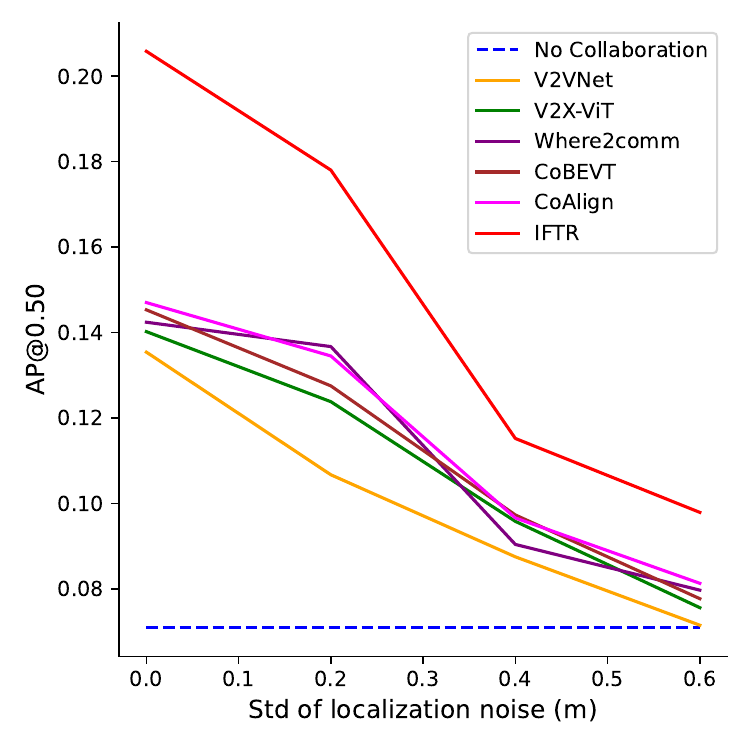}
        \caption{DAIR-V2X}
        \label{fig:noise_v2xset}
    \end{subfigure}
    \hfil 
    \begin{subfigure}{0.45\textwidth}
        \centering
        \includegraphics[width=\linewidth]{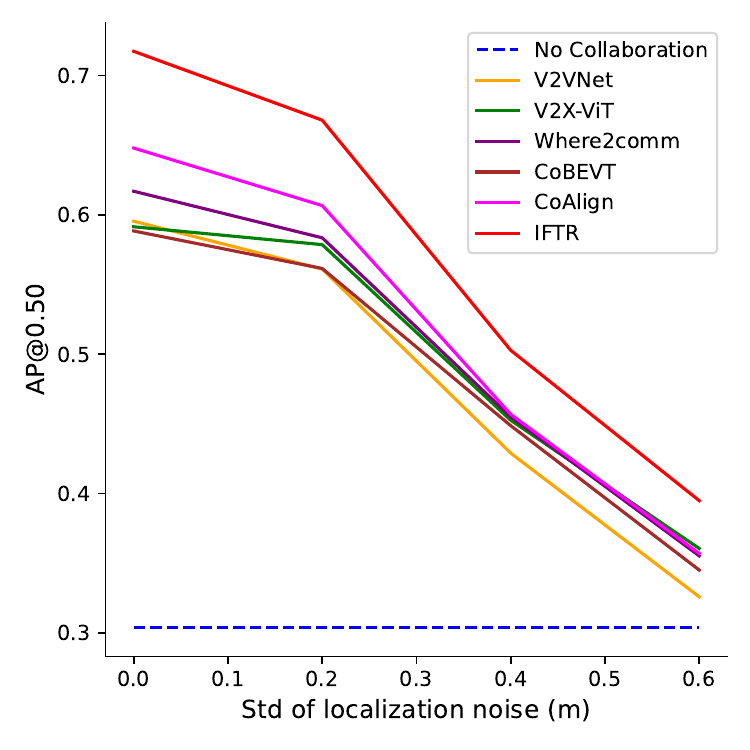}
        \caption{V2XSet}
        \label{fig:noise_dair}
    \end{subfigure}
    \caption{
    Robustness to localization noise on the DAIR-V2X and V2XSet datasets. Gaussian noise with zero mean and a varying variance is introduced. IFTR consistently outperforms previous SOTAs.
    \label{fig:noise}
    }
    
\end{figure}

\cref{tab:performance_metrics} compares the proposed IFTR with previous collaborative methods. We consider single-agent detection without collaboration (No collaboration), V2VNet \cite{wang2020v2vnet}, V2XViT \cite{xu2022v2x}, Where2comm \cite{hu2022where2comm}, CoBEVT \cite{xu2022cobevt}, CoAlign \cite{lu2023robust} and late fusion, where late fusion directly exchanges 3D detection boxes. We observe that the proposed IFTR outperforms previous methods on both real-world and simulated datasets, demonstrating the superiority of our model and its robustness to various realistic noises. Specifically, the SOTA performance of AP@70 on the V2XSet, OPV2V and DAIR-V2X datasets is improved by 12.99\%, 9.23\% and 57.96\%, respectively. Compared to previous collaboration methods at the BEV level, IFTR significantly enhances the quality of BEV features through instance feature aggregation, and adapts the cross-domain query adaption to implicitly encode the positions of the target in 3D space, achieving more accurate perception.

\subsubsection{Number of agents.}
In this experiment, we investigate the impact of the quantity of agents on the perceptual performance of IFTR on the OPV2V dataset.
As depicted in \cref{fig:agent_all}, the perceptual performance  positively increases as more agents engage in cooperative perception.
From \cref{fig:agent_certain}, it can be observed that, in a simple perception environment, the IFTR performance exceeds that of the single LiDAR-based detector \cite{lang2019pointpillars} when the collaboration count is 2 for the AP@50 metric. Similarly, for the AP@70 metric, the IFTR performance surpasses the single LiDAR-based detector when the collaboration count is 5. 
It is noteworthy that the detection performance consistently improves with the increase in the number of collaborative agents. Even upon reaching the maximum collaboration quantity in the dataset, a rising trend in performance persists. Therefore, we advocate for active collaboration among more agents to achieve increasingly advanced perceptual performance.

\subsubsection{Robustness to localization noise.}
We follow the localization noise settings in V2X-ViT \cite{xu2022v2x} (Gaussian noise with a mean of 0m and a variance of 0m-0.6m) and validate the robustness of IFTR against realistic localization noise. 
\cref{fig:noise} illustrates the detection performances as a function of localization noise level in V2XSet and DAIR-V2X datasets, respectively. We see: 
i) with the increasing localization noise, the performance of all intermediate collaborative methods deteriorates due to the misalignment of feature maps, while IFTR outperforms previous SOTAs at all the localization noise levels; 
ii) IFTR consistently outperforms the No Collaboration by a significant margin. 
The reason is the powerful transformer architecture in IFA can capture crucial perceptual cues and holistic information across agents.

\subsection{Ablation Studies}
\subsubsection{Effectiveness of different modules in IFTR.}

\begin{table}[tb]
  \caption{
  Ablation Study results of the proposed components on the OPV2V and DAIR-V2X datasets. IFA, CDQA respectively represent the incorporation of i) instance feature aggregation, and ii) cross-domain query adaptation. MASK denotes interacting exclusively with the foreground features of collaborators in the multi-view feature aggregation module.
  }
  \label{tab:combinations_results}
  \centering
  \begin{tabular}{cccc|cccc|cccc}
    \hline
    & \multirow{2}{*}{IFA} & \multirow{2}{*}{CDQA} & \multirow{2}{*}{MASK} 
    & \multicolumn{4}{c|}{OPV2V} & \multicolumn{4}{c}{DAIR-V2X} \\
    & & & & & AP@50 & AP@70 & Comm & & AP@50 & AP@70 & Comm \\ \hline
     & & & & &
    0.7762 & 0.5192 & 11.45 & & 0.1631 & 0.0421 & 12.04 \\
     & \checkmark & & & &
    0.8480 & 0.6487 & 24.23 & & 0.1998 & 0.0574 & 21.91 \\
     & \checkmark & \checkmark & & &
    0.8556 & 0.6604 & 24.23 & & 0.2058 & 0.0665 & 21.91 \\
     & \checkmark & \checkmark & \checkmark & &
    0.8458 & 0.6300 & 20.61 & & 0.1978 & 0.0618 & 18.23 \\ \hline
  \end{tabular}
\end{table}

Here we investigate the effectiveness of various components in IFTR, with the base model being BEVFormer-S incorporating a late fusion strategy. We progressively adding i) IFA, ii) CDQA, and iii) MASK to assess the impact of each component on detection performance and communication volume, where MASK refers to interacting only with the foreground features of collaborators in the MVFA module. As \cref{tab:combinations_results} demonstrates i) IFA contributes the most significant performance gain, increasing AP@70 on OPV2V and DAIR-V2X by 24.94\% and 36.34\%, respectively. This is attributed to the MVFA design in IFA, which significantly improves the quality of BEV features for individual agents, enhancing BEV features and further improving the perceptual effectiveness of late fusion; ii) CDQA increases AP@70 on OPV2V and DAIR-V2X by 1.80\% and 15.85\%, respectively. This is because CDQA encodes 2D instances into 3D object queries, making the detection head less dependent on the initially learned object queries during training (i.e., generating object queries adaptively based on 2D instances that are more suitable for 3D detection of the instance). It effectively alleviates the problem of differences in spatial distribution between the test set and the training set, and we speculate that this module will bring more significant performance improvement when there is a greater spatial distribution difference; iii) MASK reduces communication overhead by 12.30 times on the OPV2V dataset and 12.82 times on the DAIR-V2X dataset, the detection performance only decreased by 4.60\% and 7.07\% on the respective datasets for the AP@70 metric. This is highly useful for bandwidth-constrained application scenarios. It is worth noting that our proposed IFTR can also be combined with other fusion methods (such as CoBEVT \cite{xu2022cobevt} and CoAlign \cite{lu2023robust}), and we encourage researchers to explore their combined usage, believing it will lead to more advanced detection performance.

\subsubsection{Different positional encoding schemes in CDQA.}

\begin{table}[tb]
  \caption{
   Comparison of different positional encoding strategies in the CDQA module on the OPV2V and DAIR-V2X datasets. 
   None denotes not using positional encoding, Learned denotes the use of trainable positional encoding,
   and CE stands for employing the proposed cone encoding.
   }
  \label{tab:position_encoding}
  \centering
  \begin{tabular}{c|cc|cc}
    \hline
    \multirow{2}{*}{Strategy} & \multicolumn{2}{c|}{OPV2V} & \multicolumn{2}{c}{DAIR-V2X} \\
                              & AP@50        & AP@70       & AP@50         & AP@70        \\ \hline
    None                      
    & 0.8493       & 0.6532      & 0.1950        & 0.0610       \\
    Learned                   
    & 0.8532       & 0.6588      & 0.2025        & 0.0628       \\
    CE            
    & 0.8556       & 0.6604      & 0.2058        & 0.0665       \\ \hline
  \end{tabular}
\end{table}

Due to the crucial importance of positional encoding in transformer-based architectures, in this subsection, we compare cone encoding with other commonly used positional encoding methods. As can be seen from \cref{tab:position_encoding}, the performance using cone encoding is superior to other methods. This is attributed to cone encoding providing a potential cone space prior for instances, which can alleviate the difficulty of the detection head in capturing targets.

\subsection{Qualitative Evaluation}
\subsubsection{Visualization of detection results.}

\begin{figure}[tb]
  \centering
  \includegraphics[width=10cm]{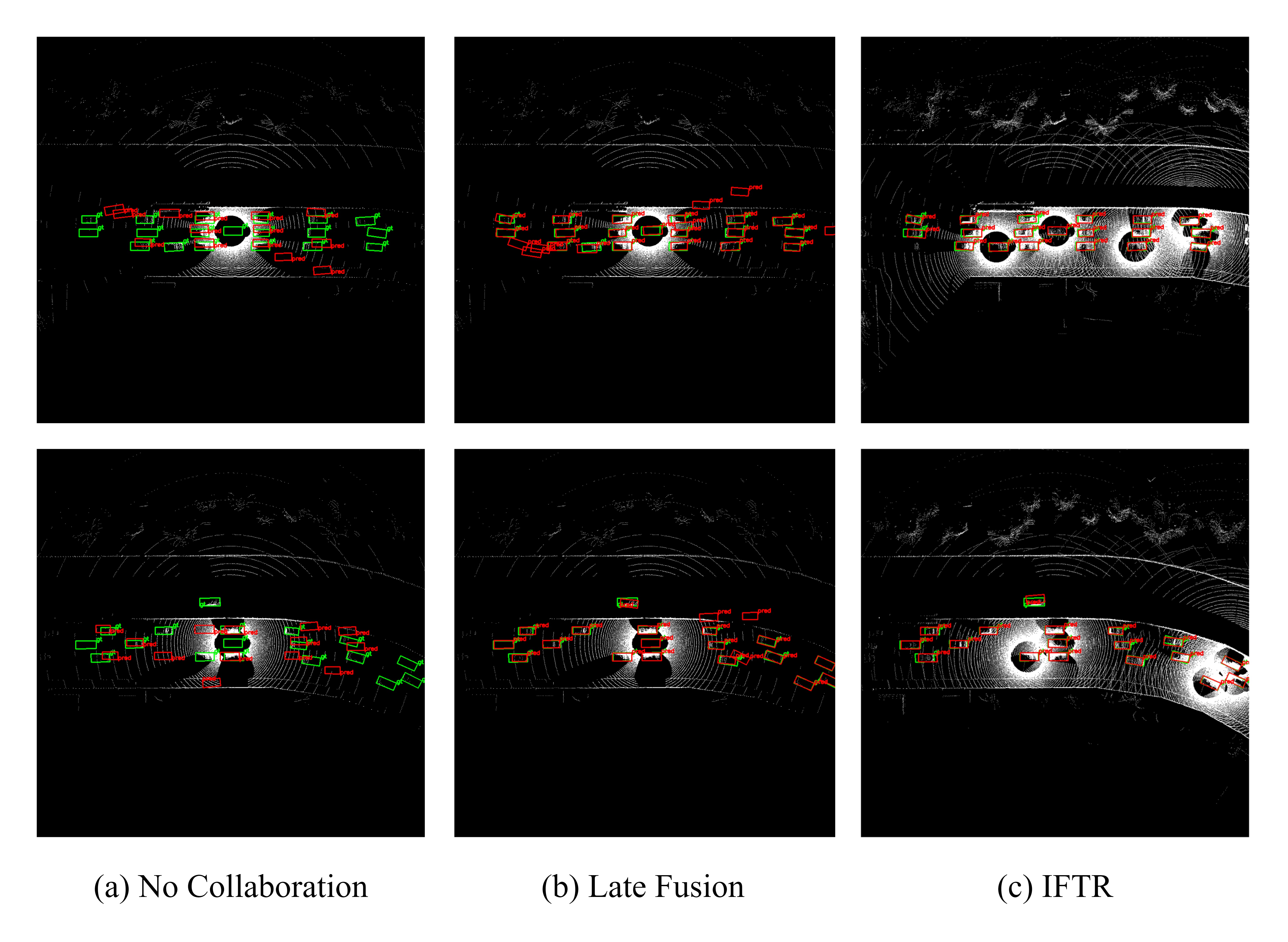}
  \caption{
  Visualization of predictions from (a) No Collaboration, (b) Late Fusion and (c) IFTR on the OPV2V test set. Green and red 3D bounding boxes represent the ground truth and prediction respectively.
  }
  \label{fig:vis_res}
\end{figure}

We conducted a qualitative analysis of the model's performance using typical samples from the OPV2V dataset, as illustrated in \cref{fig:vis_res}. The first column (a) shows the perception results of no collaboration, the second column (b) shows results with late fusion, and the third column (c) shows results using IFTR for collaborative perception. Overall, our approach generates more prediction bounding boxes that perfectly align with the ground truth. This improvement can be attributed to two main factors: i) the proposed IFA integrates meaningful information from nearby agents, significantly enhancing the quality of BEV features, and thereby improving detection performance; ii) the proposed CDQA provides effective spatial cues for objects in the scene, effectively compensating for the differences in target spatial distribution between the test set and the training set.

\section{Conclusion}
IFTR is an instance-level feature fusion framework based on transformer that aims to enhance the detection performance of camera-only collaborative perception systems through the communication and sharing of visual features. It comprises three key components: message selection and feature map reconstruction, instance feature aggregation, and cross-domain query adaptation. The message selection and feature map reconstruction module optimize communication efficiency by transmitting instance-level features. Instance feature aggregation interacts with the visual features of individual agents using predefined grid-shaped BEV queries, generating more comprehensive and accurate BEV features. The cross-domain query adaptation implicitly encodes candidate positions of targets, heuristically integrating 2D priors. Extensive experiments on several multi-agent datasets show the effectiveness of IFTR and the necessity of all its components.

\section*{Acknowledgements}
The work was supported by the Key Research \& Development
Plan of Zhejiang Province under Grant 2021C01196, in part by the Zhejiang Provincial Natural Science Foundation of China under Grant LR23F010006, in part by the Open Research Project Programme funded by the Science and Technology Development Fund (SKL-IOTSC(UM)-2024-2026) and the State Key Laboratory of Internet of Things for Smart City (University of Macau) (Ref. No.: SKL-IoTSC(UM)-2024-2026/ORP/GA01/2023).


%
%
\bibliographystyle{splncs04}
\bibliography{main}
\end{document}